# Annotation and Classification of Relevant Clauses in *Terms-and-Conditions* Contracts

**Pietro Giovanni Bizzaro[1,2], Elena Della Valentina[2], Nadia Mana[1], Maurizio Napolitano[1], and Massimo Zancanaro[1,2]**

[1]Fondazione Bruno Kessler (FBK), [2]University of Trento
{pbizzaro, mana, napolita, zancana}@fbk.eu
elena.dellavalentina@studenti.unitn.it

**Abstract**
In this paper, we propose a new annotation scheme to classify different types of clauses in *Terms-and-Conditions* contracts with the ultimate goal of supporting legal experts to quickly identify and assess problematic issues in this type of legal documents. To this end, we built a small corpus of *Terms-and-Conditions* contracts and finalized an annotation scheme of 14 categories, reaching an inter-annotator agreement of 0.92. Then, for 11 of them, we experimented with binary classification tasks using *few-shot* prompting with a multilingual T5 and two fine-tuned versions of two BERT-based LLMs for Italian. Our experiments showed the feasibility of automatic classification of our categories by reaching accuracies ranging from .79 to .95 on validation tasks.

## 1. Introduction

The application of Artificial Intelligence techniques to the Law field is a growing area of research (Bench-Capon, 2022; Trautmann et al., 2022). In particular, the research interests focus on applying machine learning techniques to various legal applications (Villata et al., 2022). Recently, the power of Large Language Models (LLMs) as foundational models has started to be leveraged to automatically analyze legal documents (Choi, 2023).

In this paper, we propose a new annotation scheme to classify different types of clauses in *Terms-and-Conditions* contracts and provide a baseline for the automatic classification of clauses.

*Terms-and-Conditions* contracts are characterized by their consistency and comprehensiveness, crafted to cater to a diverse audience and varied legal and judicial contexts. These types of contracts, which have spread widely with the advent of online services, embody a rich source of legal and technical nuances that are often complex and subtle, requiring careful and thorough analysis to assess the legal implications (Braun and Matthes, 2020).

Our goal is to eventually support legal experts in identifying and assessing problematic issues in this type of legal document rather than replacing their work. In this respect, the categories used for the annotation and classification of relevant clauses have been explicitly designed with the aim of helping experts to spot relevant parts of a contract quickly.

Our work is similar to (Lippi et al. 2019). However, Lippi and colleagues aimed to empower consumers by detecting unfair terms. Therefore, they focused on five categories of potentially unfair clauses identified by Loos and Luzak (2016).

We started from the more extensive set of categories proposed by the Atticus Project (Hendrycks et al., 2021), and we selected the most relevant ones for the specific *Terms-and-Conditions* type of contracts. We identified and refined 33 labels, and manually clustered them into 14 categories. Then, we annotated a corpus of 3626 clauses, from 41 contracts written in Italian.

For the preliminary work presented in this paper, we limited our analysis to the 11 most frequent categories in the corpus. We experimented with *few-shot* prompting and fine-tuning of two LLMs to provide a baseline for automatic classification.

## 2. Related work

Yu et al. (2023) delved into the effectiveness of zero-shot and few-shot learning, especially when combined with reason-based prompting mechanisms, in interpreting generic legal texts taken from the Japanese bar exam, where aspiring lawyers are required to determine whether a given legal statement is true or false.

In the context of contract understanding, our study is methodologically similar to Elwany et al. (2019), who demonstrated the efficacy of fine-tuning a BERT model (Devlin et al., 2019) by refining it with an annotated corpus of legal agreements.

The present study also resonates with the work by Licari and Comandè (2022), where BERT models applied within the Italian legal context are exploited to interpret and process legal texts using corpora of civil law judgments.

Sarti and Nissim (2022) marked the introduction of IT5, a model pre-trained specifically on the Italian language and consistently able to outperform its multilingual counterparts on Italian language tasks, providing the best scale-to-performance ratio across tested models. Their focus is not confined to the legal sphere, but it spans various applications, showcasing the versatility of IT5 in outclassing multilingual counterparts in terms of performance on Italian-specific tasks.

Lastly, the CLAUDETTE project (Lippi et al., 2019) made significant strides in identifying unfair terms in consumer contracts, especially online platform terms and conditions. In particular, the authors stated that the system can automatically detect over 80% of potentially unfair clauses, with an 80% precision. Our work has a similar but different potential target users, and a larger set of categories. On the other hand, we are not interested in detecting unfair clauses but rather to categorize the typology of clauses.

The present study focuses on the Italian language, extending the reach of automated contract review beyond English and aiming to overcome language disparity in text classification tasks (Verma et al., 2022). Our work seeks to take a further step in diversifying the approaches in legal text analysis. Our focus on a broader array of contractual elements and the inclusion of Italian language contracts highlights a complementary perspective with respect to the related work above.

## 3. Methodology

For developing our coding scheme, we started from the model proposed in the Atticus Project (Hendrycks et al., 2021) with appropriate adjustments to make it suitable for our application domain (*Terms-and-Conditions* contracts) and the European civil law system.

### 3.1 Annotation process

The annotation process was done by two authors with legal expertise in European and Italian contractual law.

The preliminary phase consisted of selecting labels and categories, mostly identified in and refined from the Atticus project, as well as in formulating a first batch of annotation rules. In particular, the Contract Understanding Atticus Dataset (CUAD) comprises 41 labels referring to commercial contracts governing the relationship between corporations. Within the scope of goals, we identified 25 of those labels that apply to the analysis of *Terms-and-Conditions* contracts and discarded those that were (a) pertaining specifically to the commercial domain, (b) peculiar to the American (common law) legal tradition and (c) redundant within other labels. Furthermore, 8 additional labels have been included. These were specifically pertinent for the European civil law system and the Italian *Terms-and-Conditions* domain and not present in CUAD.

To build a functional kit for automated sentence classification tasks, the 33 labels of our coding scheme have been manually clustered into 14 categories, covering the main legal aspects characterizing the *Terms-and-Conditions contracts*. For example, the three labels *party*, *change of control*, and *third beneficiary,* originally identified in CUAD, are considered part of the overall *party* category. Additionally, a residual category has been included to annotate text segments deemed "not relevant", i.e. those that refer to provisions that do not create or amend the *subjective rights* (Dedek, 2010) of a subject; therefore, there are no legal consequences associated with the textual wording.

Firstly, one contract was annotated by both annotators and then discussed to reconcile the different interpretations and refine the annotation rules. Then, 30 contracts were consistently annotated by the two experts. In addition, 10 other contracts were partially annotated with specific attention to the less frequent categories in the corpus. Ultimately, 2815 clauses were annotated with 14 categories, while 811 as "not relevant", for a total of 3626 clauses.

Finally, an inter-coder agreement was also performed to assess the coding scheme. The inter-coder agreement was computed on 20% of the clauses from the last 10 contracts. Cohen's kappa on these annotations is 0.92, which may be considered an excellent agreement according to Fleiss (1981).

However, there are a few discrepancies that are crucial to examine for refining the annotation process: the "*warranty*" category has been confused in two cases, respectively, with "*party*" and "*liability*". That could imply some overlap or ambiguity in the definitions of these categories due to legal considerations. Indeed, the "*warranty*" category refers to clauses establishing or excluding a warranty against defects or errors. That is often linked to forms of liability of the producer or service provider, hence to the "*liability*" category. The confusion in the case of "*warranty*" and "*party*" is instead linked to the peculiarity of the annotated clause: the clause establishes indeed both a form of warranty towards one of the parties ("*warranty*" category) and the possibility for third parties to be involved (*"party"* category). The difference in the categories assigned to the clause depends on the relevance each annotator gave to the content of the clause. Also, "*remedy*" and "*liability*" have led to confusion that might stem from the close relationship between these two categories, as remedies are often sought for liabilities. Several discussions and examples are reported in the annotation scheme[1] to assist the annotators in minimizing this confusion.

The discrepancies observed, though minimal, were instrumental for the iterative refinement of the annotation guidelines by providing clearer definitions, edge-cases, and examples. Further discussion among the annotators can also be facilitated to address the mentioned areas of discrepancy, ensuring that the coding scheme becomes even more reliable for larger-scale annotation tasks.

### 3.2 Selected categories

A brief description of the adopted 14 categories is provided below:

- *Acceptance* refers to clauses that establish and manage how the contract is accepted.
- *Amendment* includes both clauses regulating unilateral amendment of the contract and clauses that allow the

---
[1] https://github.com/i3-fbk/LLM-PE_Terms_and_Conditions_Contracts

- modification of the performance that the contract refers to.
- *Competence* includes information explaining aspects related to either applicability (concerning applicable law or hierarchies between different contracts signed by the parties) or to the competence of courts or means of alternative dispute resolution.
- *Data protection* defines the usage, rights, and obligations on processing of personal data of the parties and third parties.
- *Date* refers to relevant dates in the contract, such as when the contract is signed, when it comes into force, and when the contract expires or is renewed.
- *Intellectual Property* includes clauses assigning or managing rights referring to the intellectual property of a product, service, or any other element the contract manages.
- *License* includes clauses establishing the concession of a license, its management, and limitations.
- *Liability* refers to clauses establishing liability allocation, exclusion, or limitation.
- *Object* indicates the good or the service that is the object of the contract.
- *Party* includes information about the parties who signed the contract, including their names, characteristics, rules on third beneficiaries, and party change.
- *Remedy* refers to clauses establishing extracontractual solutions, such as orders for payment or penalties, to remedy a contract violation.
- *Term* refers to the contract's relevant terms, such as renewal or forewarning terms, terms related to termination clauses, or terms establishing license durations.
- *Termination* refers to clauses containing information or establishing rules in relation to the possibility for either party to terminate the contract unilaterally.
- *Warranty* defines an explicit guarantee of performance from one party to the other.

## 4. The classification exercises

To produce a first baseline, we initially tried a few-shot prompting approach. As discussed by Choi (2023), thanks to the current level of LLMs maturity, good results might be reached even with simpler methods. Then, we fine-tuned two LLM models for Italian. Similarly to Song et al. (2022), all the classifications were binary with train and validation datasets balanced with respect to positive and negative examples. Differently from Choi (2023), we decided to use Open Source LLM models rather than proprietary ones.

For the classification exercise, we considered the 11 categories among the ones described above (see 3.2) with at least 70 positive instances. For each category, we created a training dataset containing 70% of the positive instances and a validation dataset with the remaining data entries. The negative instances were randomly sampled from the rest of the corpus to balance the positive instances in the datasets.

The sizes of the training datasets vary between 1,157 instances ( positive and negative) for the *warranty* category to 98 for the *term* category (see Table 1). For the validation datasets, the sizes vary from 497 for *warranty* to 42 for *term* (see Table 2).

### 4.1 The few-shot prompting classification

For the *few-shot* prompting classification, we used Google's FLAN-T5 large (Chung et al., 2022), a multilingual version of T5 (Raffel et al., 2020) fine-tuned with instructions.

For each instance, one positive and one negative example was chosen from the training dataset and used to build the prompt. Each clause in the validation dataset was then classified using the prompt. For this task, the training dataset was not used.

| Category | Base_Model | Train_Size | Accuracy | AUC | Loss |
|---|---|---|---|---|---|
| acceptance | Italian BERT XXL | 127 | 0.88 | 0.89 | 0.4099 |
| | **ITALIAN-LEGAL-BERT** | | **0.92** | **0.92** | **0.3304** |
| amendment | Italian BERT XXL | 207 | 0.76 | 0.87 | 0.5070 |
| | **ITALIAN-LEGAL-BERT** | | **0.91** | **0.98** | **0.2030** |
| competence | **Italian BERT XXL** | 485 | **0.95** | 0.99 | **0.1300** |
| | ITALIAN-LEGAL-BERT | | 0.93 | 1.00 | 0.1270 |
| data protection | **Italian BERT XXL** | 194 | **0.80** | **0.80** | **0.4060** |
| | ITALIAN-LEGAL-BERT | | 0.75 | 0.08 | 0.5090 |
| intellectual property | **Italian BERT XXL** | 130 | **0.82** | **0.95** | **0.3490** |
| | ITALIAN-LEGAL-BERT | | 0.82 | 0.90 | 0.4150 |
| liability | Italian BERT XXL | 484 | 0.95 | 0.99 | 0.1270 |
| | **ITALIAN-LEGAL-BERT** | | **0.96** | **1.00** | **0.1150** |
| license | **Italian BERT XXL** | 362 | 0.90 | **0.97** | **0.2370** |
| | ITALIAN-LEGAL-BERT | | **0.93** | 0.96 | 0.2440 |
| party | Italian BERT XXL | 295 | 0.83 | 0.92 | 0.3700 |
| | **ITALIAN-LEGAL-BERT** | | **0.92** | **0.95** | **0.2910** |
| term | Italian BERT XXL | 98 | 0.76 | 0.87 | 0.4650 |
| | **ITALIAN-LEGAL-BERT** | | **1.00** | **1.00** | **0.0100** |
| termination | Italian BERT XXL | 266 | 0.82 | 0.93 | 0.4040 |
| | **ITALIAN-LEGAL-BERT** | | **0.91** | **0.94** | **0.3150** |
| warranty | **Italian BERT XXL** | 1157 | **0.81** | **0.91** | **0.4210** |
| | ITALIAN-LEGAL-BERT | | 0.79 | 0.89 | 0.4540 |

Table 1: Figure of merits of the classification exercises with the *fine-tuning* of the two models, Italian BERT XXL and ITALIAN-LEGAL-BERT (best models and results are in bold)

The results are summarized in Table 2. Overall, the *few-shot* prompting classification is not satisfactory. Although a couple of categories (*acceptance* and *intellectual property*) reach a reasonable accuracy (0.82 and .70, respectively), all the others barely reach the level of chance. Of course, it might be that a more considerate choice of the examples for the prompt (rather than random sampling) would improve the classification.

It is worth noting that the LLMs used a label different than the classifying category or the class *other* in several cases. For example, for *intellectual property*, the classifier sometimes proposed "trademark", "intellectual", "user", "privacy". Since *few-shot* prompting is a generative approach (albeit used here as a classifier), this phenomenon may be considered a case of a hallucination (Ji et al., 2022). The impact of these hallucinations ranged from 0 for the category *acceptance* to 42% of the target classification for *intellectual property*. Nevertheless, since we trained binary classifiers, we could easily fix this issue by taking any output of the classifier different from *other* as the target category (the validation results reported in Table 2 reflect this approach), yet that might represent a problem in the case of multi-class classifications.

## 4.2 Fine tuning of BERT-based models

To improve the baseline, we fine-tuned two BERT-based models (Devlin et al., 2019) pre-trained on Italian for a textual classification task, namely the general-purpose Italian BERT XXL Cased[2] and its extension ITALIAN-LEGAL-BERT (Licari & Comandè, 2022) with additional pre-training of the Italian BERT model on Italian civil and criminal law corpora.

For the fine-tuning exercise, we used the datasets discussed above. We built 11 binary classifiers for each of the two BERT-based models. Each classifier was built by fine-tuning one of the two models using the respective training dataset. The actual training was performed using the AutoTrain online service by HuggingFace[3].

The fine-tuning process leverages the transformer architecture of the BERT models, thereby ensuring that the models' weights are calibrated to recognize the annotated legal documents.

The assessment of each model was performed by measuring the classification accuracy on the same validation datasets used in the prompting classification above.

The validation results are reported in Table 1. Overall, the classification is satisfactory, with accuracies ranging from .75 to .96. There are some differences between the two models for several categories, but the performances are quite similar.

## 5. Discussion

We excluded from the classification exercised three categories ("*audit right*," "*remedy*", and "*date*") because they do not appear frequently in our corpus. This scarcity likely stems from the limited use of the related provisions in the type of contract assessed in the present work, specifically *Terms and Conditions*. Nevertheless, it might be important to assess their relevance in the context: on one side, the inclusion of "*audit rights*" and the provision of additional non-judicial "*remedies*" within consumer contracts are not standard practices, but on the other side, precisely because of their lower use, it might interesting to make the user aware of their presence. Additionally, identifying specific "*dates*" within *Terms and Conditions* contract can be deemed unnecessary in agreements where the conclusive and termination actions often align with engaging with the services or platform.

For the validation stage, the performance metrics generally suggest high accuracy for both models. Notably, the ITALIAN-LEGAL-BERT model exhibited marginally better performance than the Italian BERT XXL, and in some cases the general-purpose model even slightly outperformed its domain-specific counterpart. Although seemingly counterintuitive, similar observations were made by Licari and Comandè (2022) in their study, where the fine-tuned model underperformed compared to the general-purpose tasks reliant on trained criminal cases law. We think that it might be the case that the knowledge required to detect our categories lies more in the language patterns than in the specific legal linguistic knowledge. However, full understanding of this effect requires further investigation and an expanded corpus to facilitate a more comprehensive analysis.

Finally, we note that the few-shot prompting strategy yielded surprisingly unsatisfactory results, considering previous results presented in literature (Yu et al., 2023). However, exploring alternative strategies for selecting the prompt and possibly testing different Language Model Models (LLMs) may lead to improvement.

## 6. Conclusion

The primary purpose of this work was to validate a coding scheme created for the goal of automatically tagging the different types of clauses in Italian *Terms-and-Condition* contracts.

We started from the categorization proposed in the Atticus Project and filtered and adapted the categories to best fit our case study and the characteristics of the Italian legal system.

The coding scheme demonstrated good reliability. We then annotated a small, but consistent, annotated corpus. Three automatic classification exercises were conducted to assess baselines for further improvements.

Our ultimate objective is to provide legal experts with a tool to support a more thorough and efficient contract review process. Although the results are still preliminary, we believe that this work can contribute to the general topic of AI and Law, and specifically focus the research on concrete applications.

Future research directions may include expanding the annotated corpus to cover better all the categories, as well as to include other legal domains for a more comprehensive analysis. Furthermore, exploring alternative strategies for few-shot prompting and testing different model architectures could improve automatic annotation accuracy. Finally, additional comparative studies between domain-specific and general-purpose models will help identify effective approaches for handling legal text.

## 7. Acknowledgements

This research received partial support from the PNRR project FAIR-Future AI Research

---

[2] https://huggingface.co/dbmdz/bert-base-italian-xxl-cased
[3] https://huggingface.co/autotrain

(PE00000013), under the NRRP MUR program funded by the NextGenerationEU.

| Category | Base_Model | Val_Size | True pos | False pos | True nega | False negatives | Accuracy | Precision | Recall |
|---|---|---|---|---|---|---|---|---|---|
| acceptance | FS prompting | 55 | 20 | 9 | 25 | 1 | 0.82 | 0.69 | 0.95 |
| | Italian BERT XXL | | 26 | 3 | 20 | 6 | 0.84 | **0.90** | 0.81 |
| | **ITALIAN-LEGAL-BERT** | | 26 | 3 | 21 | 5 | **0.85** | **0.90** | **0.84** |
| amendment | FS prompting | 89 | 19 | 31 | 34 | 5 | 0.60 | 0.38 | 0.79 |
| | Italian BERT XXL | | 40 | 10 | 30 | 9 | 0.79 | 0.80 | 0.82 |
| | **ITALIAN-LEGAL-BERT** | | 44 | 6 | 32 | 7 | **0.85** | **0.88** | **0.86** |
| competence | FS prompting | 209 | 33 | 82 | 82 | 12 | 0.55 | 0.29 | 0.73 |
| | **Italian BERT XXL** | | 105 | 10 | 93 | 1 | **0.95** | **0.91** | **0.99** |
| | ITALIAN-LEGAL-BERT | | 104 | 11 | 90 | 4 | 0.93 | 0.90 | 0.96 |
| data protection | FS prompting | 84 | 11 | 31 | 41 | 1 | 0.62 | 0.26 | 0.92 |
| | **Italian BERT XXL** | | 36 | 6 | 38 | 4 | **0.88** | **0.86** | **0.90** |
| | ITALIAN-LEGAL-BERT | | 36 | 6 | 34 | 8 | 0.83 | **0.86** | 0.82 |
| intellectual property | FS prompting | 56 | 10 | 16 | 29 | 1 | 0.70 | 0.38 | 0.91 |
| | **Italian BERT XXL** | | 24 | 2 | 26 | 4 | **0.89** | **0.92** | **0.86** |
| | ITALIAN-LEGAL-BERT | | 24 | 2 | 24 | 6 | 0.86 | **0.92** | 0.80 |
| liability | FS prompting | 208 | 29 | 84 | 86 | 9 | 0.55 | 0.26 | 0.76 |
| | **Italian BERT XXL** | | 101 | 12 | 84 | 11 | **0.89** | 0.89 | **0.90** |
| | ITALIAN-LEGAL-BERT | | 102 | 11 | 82 | 13 | 0.88 | **0.90** | 0.89 |
| license | FS prompting | 156 | 24 | 57 | 69 | 6 | 0.60 | 0.30 | 0.80 |
| | Italian BERT XXL | | 70 | 11 | 60 | 15 | **0.83** | 0.86 | **0.82** |
| | **ITALIAN-LEGAL-BERT** | | 77 | 4 | 52 | 23 | **0.83** | **0.95** | 0.77 |
| party | FS prompting | 127 | 11 | 58 | 55 | 3 | 0.52 | 0.16 | 0.79 |
| | Italian BERT XXL | | 61 | 8 | 48 | 10 | 0.86 | 0.88 | **0.86** |
| | **ITALIAN-LEGAL-BERT** | | 65 | 4 | 47 | 11 | **0.88** | **0.94** | **0.86** |
| term | FS prompting | 42 | 8 | 15 | 18 | 1 | 0.62 | 0.35 | 0.89 |
| | **Italian BERT XXL** | | 19 | 4 | 19 | 0 | 0.90 | 0.83 | **1.00** |
| | **ITALIAN-LEGAL-BERT** | | 22 | 1 | 18 | 1 | **0.95** | **0.96** | 0.96 |
| termination | FS prompting | 114 | 3 | 54 | 54 | 3 | 0.50 | 0.05 | 0.50 |
| | **Italian BERT XXL** | | 53 | 4 | 43 | 14 | 0.84 | **0.93** | 0.79 |
| | **ITALIAN-LEGAL-BERT** | | 52 | 5 | 48 | 9 | **0.88** | 0.91 | **0.85** |
| warranty | FS prompting | 497 | 3 | 244 | 238 | 12 | 0.48 | 0.01 | 0.20 |
| | **Italian BERT XXL** | | 222 | 25 | 181 | 69 | **0.81** | **0.90** | 0.76 |
| | **ITALIAN-LEGAL-BERT** | | 209 | 38 | 193 | 57 | **0.81** | 0.85 | **0.79** |

Table 2: Validation results of the classification exercises on 11 categories using *few-shot (FS) prompting* and Italian BERT XXL and ITALIAN-LEGAL-BERT models (best models and results are in bold).

## 8. Bibliographical references

Bench-Capon, T. (2022). Thirty years of Artificial Intelligence and Law: Editor's Introduction. In *Artificial Intelligence and Law*, 30(4), 475-479. https://doi.org/10.1007/s10506-022-09325-8

Braun, D. and Matthes, F. (2020). Automatic Detection of Terms and Conditions in German and English Online Shops. In *Proceedings of the 16th International Conference on Web Information Systems and Technologies-WEBIST*.

Choi, J. H. (2023). How to Use Large Language Models for Empirical Legal Research. In *Journal of Institutional and Theoretical Economics* (Forthcoming), Minnesota Legal Studies Research Paper No. 23-23, Available at SSRN: https://ssrn.com/abstract=4536852

Chung, H. W., Hou, L., Longpre, S., Zoph, B., Tay, Y., Fedus, W., Li, Y., Wang, X., Dehghani, M., Brahma, S., Webson, A., Gu, S. S., Dai, Z., Suzgun, M., Chen, X., Chowdhery, A., Castro-Ros, A., Pellat, M., Robinson, K., Wei, J. (2022). Scaling Instruction-Finetuned Language Models. ArXiv, abs/2210.11416. http://arxiv.org/abs/2210.11416

Dedek, H. (2010). From Norms to Facts: The Realization of Rights in Common and Civil Private Law. *McGill Law Journal / Revue de droit de McGill*, 56(1), 77-114. https://doi.org/10.7202/045699ar

Devlin, J., Chang, M.-W., Lee, K., & Toutanova, K. (2019). BERT: Pre-training of Deep Bidirectional Transformers for Language Understanding. In *Proceedings of the 2019 Conference of the North*

American Chapter of the Association for Computational Linguistics: Human Language Technologies* (Volume 1), 4171-4186.

Elwany, E., Moore, D.A., & Oberoi, G. (2019). BERT Goes to Law School: Quantifying the Competitive Advantage of Access to Large Legal Corpora in Contract Understanding. ArXiv, abs/1911.00473.

Fleiss, J.L. (1981). Statistical Methods for Rates and Proportions. John Wiley & Sons, 2nd Edition.

Hendrycks, D., Burns, C., Chen, A., & Ball, S. (2021). CUAD: An Expert-Annotated NLP Dataset for Legal Contract Review. ArXiv, abs/2103.06268.

Ji, Z., Lee, N., Frieske, R., Yu, T., Su, D., Xu, Y., Ishii, E., Bang, Y. J., Madotto, A., & Fung, P. (2023). Survey of Hallucination in Natural Language Generation. In *ACM Computing Surveys*, 55(12), 1-38. https://doi.org/10.1145/3571730

Lagioia, F., Jablonowska, A., Liepina, R., & Drazewski, K. (2022). AI in Search of Unfairness in Consumer Contracts: The Terms of Service Landscape. *Journal of Consumer Policy*, 45, 481-536.

Licari, D., & Comandè, G. (2022). ITALIAN-LEGAL-BERT: A Pre-trained Transformer Language Model for Italian Law. In *Proceedings of EKAW'22: Companion Proceedings of the 23rd International Conference on Knowledge Engineering and Knowledge Management*.

Lippi, M., Pałka, P., Contissa, G., Lagioia, F., Micklitz, H., W., Sartor, G., and Torroni, P. (2019). CLAUDETTE: An automated detector of potentially unfair clauses in online terms of service. In *Artificial Intelligence and Law*, 27(2), 117-139

Loos, M.B., & Luzak, J. (2015). Wanted: a Bigger Stick. On Unfair Terms in Consumer Contracts with Online Service Providers. Journal of Consumer Policy, 39, 63 - 90.

Raffel, C., Shazeer, N.M., Roberts, A., Lee, K., Narang, S., Matena, M., Zhou, Y., Li, W., and Liu, P.J. (2019). Exploring the Limits of Transfer Learning with a Unified Text-to-Text Transformer. J. Mach. Learn. Res., 21, 140:1-140:67.

Sarti, G., Nissim, M. (2022). IT5: Large-scale Text-to-text Pretraining for Italian Language Understanding and Generation. arXiv, abs/2203.03759.

Song, D., Gao, S., He, B., & Schilder, F. (2022). On the effectiveness of pre-trained language models for legal natural language processing: An empirical study. *IEEE Access*, *10*, 75835-75858.

Trautmann, D., Petrova, A., and Schilder, F. (2022). Legal prompt engineering for multilingual legal judgment prediction. arXiv, abs/2212.02199.

Verma, G., Mujumdar, R., Wang, Z. J., De Choudhury, M., & Kumar, S. (2022). Overcoming Language Disparity in Online Content Classification with Multimodal Learning. ArXiv, abs/2205.09744. https://doi.org/10.48550/ARXIV.2205.09744

Villata, S., Araszkiewicz, M., Ashley, K., Bench-Capon, T., Branting, L. K., Conrad, J. G., & Wyner, A. (2022). Thirty years of artificial intelligence and law: the third decade. In *Artificial Intelligence and Law*, 30(4), 561-591. https://doi.org/10.1007/s10506-022-09327-6

Yu, F., Quartey, L., & Schilder, F. (2023). Exploring the Effectiveness of Prompt Engineering for Legal Reasoning Tasks. In *Findings of the Association for Computational Linguistics: ACL 2023*, 13582-13596.